\documentclass[journal]{IEEEtran}
\usepackage[utf8]{inputenc}
\usepackage[fleqn]{amsmath}
\usepackage{graphicx}
\usepackage[ruled]{algorithm}
\usepackage[noend]{algpseudocode}
\usepackage{latexsym}
\usepackage{cite}

\usepackage{cite}

\ifCLASSINFOpdf
  \DeclareGraphicsExtensions{.pdf,.jpeg,.png}
\else
\fi
\ifCLASSOPTIONcompsoc
 \usepackage[caption=false,font=normalsize,labelfont=sf,textfont=sf]{subfig}
\else
  \usepackage[caption=false,font=footnotesize]{subfig}
\fi
\begin{document}
%
\title{Attention Tree: Learning Hierarchies of Visual Features for Large-Scale Image Recognition}
%
%
%

\author{Priyadarshini~Panda*, \textit{ Student Member}, \textit{IEEE},
        and Kaushik~Roy,~\IEEEmembership{Fellow,~IEEE}
\thanks{* Correspondance e-mail: pandap@purdue.edu. The manuscript is under review in IEEE Transactions on Neural Networks and Learning Systems.}
}

%
%

\markboth{Journal of \LaTeX\ Class Files,~Vol.~, No.~, July~2016}%
{Shell \MakeLowercase{\textit{et al.}}: Bare Demo of IEEEtran.cls for IEEE Journals}
%



\maketitle

\begin{abstract}
One of the key challenges in machine learning is to design a computationally efficient multi-class classifier while maintaining the output accuracy and performance. In this paper, we present a tree-based classifier: Attention Tree (ATree) for large-scale image classification that uses recursive Adaboost training \cite{collins2002logistic} to construct a visual attention hierarchy. The proposed attention model is inspired from the biological “selective tuning mechanism for cortical visual processing”. We exploit the inherent feature similarity across images in datasets to identify the input variability and use recursive optimization procedure, to determine data partitioning at each node, thereby, learning the attention hierarchy. A set of binary classifiers is organized on top of the learnt hierarchy to minimize the overall test-time complexity. The attention model maximizes the margins for the binary classifiers for optimal decision boundary modelling, leading to better performance at minimal complexity. The proposed framework has been evaluated on both Caltech-256 and SUN datasets and achieves accuracy improvement over state-of-the-art tree-based methods at significantly lower computational cost. 
\end{abstract}

%

\section{Introduction}
``Attention'' mechanisms are a critical component to brain's cognitive performance. Such mechanisms enable the brain to process overwhelming visual stimuli with limited capacity by selectively enhancing the information relevant to one's current behaviour \cite{buschman2015behavior}. 
With the massive growth of digital image data due to social media, surveillance camera, 
among others, there is a growing demand for computing platforms to perform cognitive tasks. Most of these computing platforms have limited resources in terms of processing power and battery life. Hence, researchers have been strongly motivated to design efficient large-scale image recognition methods to enable resource constrained IoT (Internet of Things) devices with cognitive intelligence \cite{fei2006one,xiao2010sun}.

Several brain-inspired computing models including Support Vector Machines (SVM) \cite{geebelen2012reducing, cortes1995support}, random forest \cite{breiman2001random}, and Adaboost \cite{yuksel2012twenty,collins2002logistic} have proven to be very successful for image recognition. However, these classifiers do not scale well with increasing number of image categories. Deep Learning Networks like ConvNets \cite{krizhevsky2012imagenet} have achieved state-of-the-art accuracies, even surpassing human performance \cite{he2015delving} for Imagenet dataset. However, they have been criticized for their enormous training cost and computational complexity. Similarly, the one-versus-all linear SVM, one of the most popular classifiers for large-scale classification, is computationally inefficient as its complexity increases linearly with the number of categories. While these classifiers are modeled to mimic the brain-like cognitive abilities, they lack the remarkable energy-efficient processing capability of the brain. The brain carries out enormously diverse and complex information processing to deal with a constantly varying world at a power budget of about ~12-20 W \cite{balasubramanian2015heterogeneity}. Seeking to attain the brain’s efficiency, we draw inspiration from its underlying processing mechanisms to design a multi-class classification method that is both accurate and computationally efficient. 

One such mechanism known as ``saliency based selective attention'' shown in Fig. 1 (left) simplifies complex visual tasks into characteristic features and then selectively activates particular areas of the brain based on the feature information in the input \cite{whitney2009neuroscience}. When presented with new visual images, the brain associates the already learnt features to the visual appearance of the new object types to perform recognition \cite{ungerleider2000mechanisms,buschman2015behavior}. This facilitates the brain to learn a host of new information with limited capacity and also speeds up the recognition process. Interestingly, we note that there is significant similarity among underlying characteristic features (like color or texture) of images across multiple objects in real world applications. This presents us with an opportunity to build an efficient visual recognition system incorporating inter-class feature similarities and relationships.

In this work, we propose a computationally efficient multi-class classification method: Attention Tree (ATree) that exploits the feature similarity among multiple classes in the dataset to build a hierarchical tree structure composed of binary classifiers. The resultant ATree learns a hierarchy of features that transition from general to specific as we go deeper into the tree in a top-down manner. This is similar to the state-of-the-art Deep Learning convolutional Networks (DLNs) where the convolutional layers exhibit a generic-to-specific transition in the learnt features \cite{yosinski2014transferable}. In case of DLNs, the entire network is utilized for the recognition of a particular test input. In contrast, the construction of the attention tree incorporates effective and active pruning of the dataset during training of the individual tree nodes resulting in an efficient instance-specific classification path. In addition, as we will see in later sections, our attention model captures both inter and intra class feature similarity to build a tree hierarchy with decision paths of varying lengths even for the same class. This provides substantial benefits in test speed and computational efficiency for large-scale problems while maintaining competitive classification accuracy.

\begin{figure}[!t]
\centering
\includegraphics[width=0.5\textwidth]{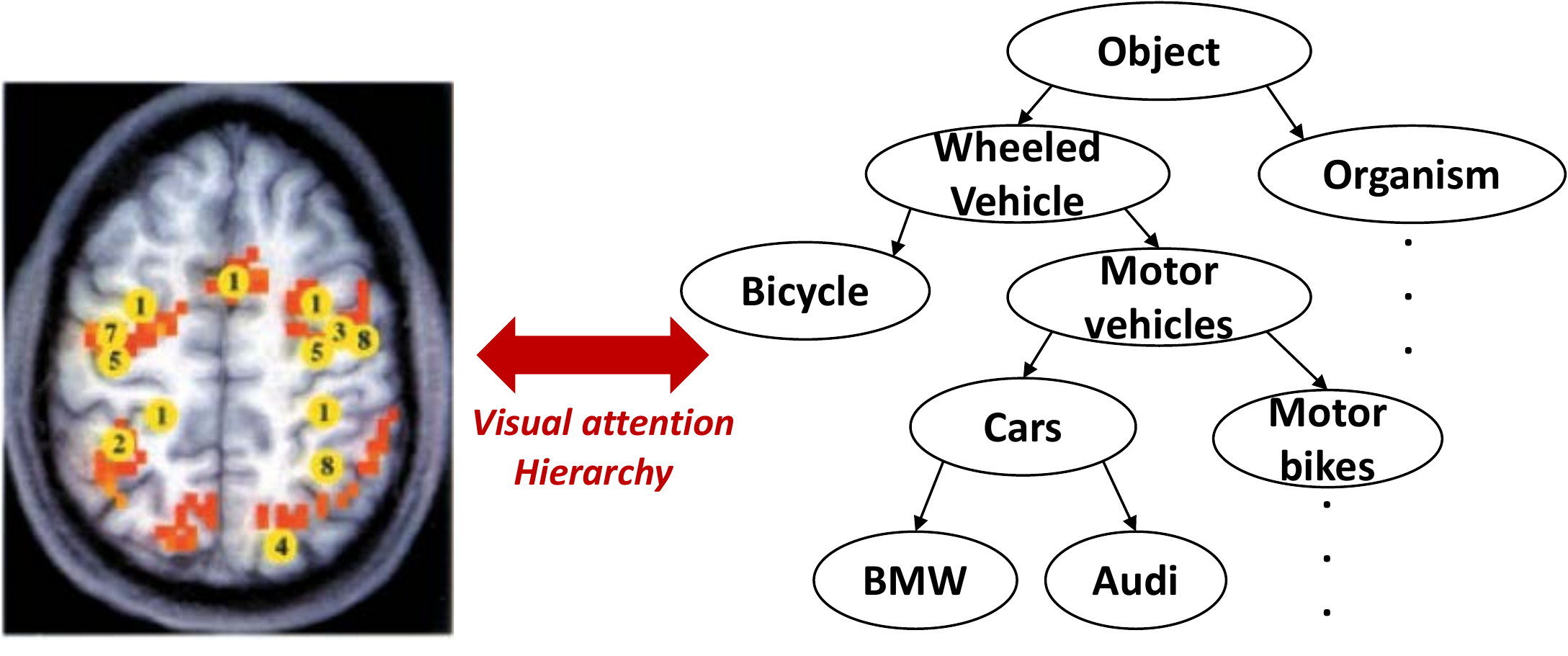}
\vspace{-4mm}
\caption{(Left)  Selective attention mechanism observed in the frontal and parietal cortex involved in the generation and control of salient attentional signals \cite{ungerleider2000mechanisms} (Right) Toy example of Attention Tree with hierarchies formed on the basis of different semantic categories and inter-class relationships of object types }
\vspace{-4mm}
\end{figure}

Fig. 1 (right) shows a toy example of an ATree based on real-world broad semantic categories for different object classes. For example, to recognise a car, it is not sensible to learn all the specific appearance details. Instead, first we learn the general vehicle-type features (wheels, shape etc) and then learn more discriminative details ( brand symbol). Thus, we learn a hierarchy of features generalizing over object instances like: Wheeled vehicle$\rightarrow$Motor vehicles$\rightarrow$Cars$\rightarrow$BMW. If presented with new motorbike object types, the attention hierarchy now associates this new category of objects to the already learnt "Wheeled vehicle" features and then learns more discriminative details corresponding to the motorbike types. Each node of the ATree is then associated with different features based on inter-class relationships. It is evident from Fig. 1 that the attention tree method bears resemblance to the selective attention mechanism of the brain (Fig. 1 left) by exploiting feature similarity and the implicit relationships among different visual data to learn a meaningful hierarchy for recognition.

\section{Related Work}

While decision tree \cite{gao2011discriminative}, ensemble methods \cite{sun2013find,he2012ssc,allwein2000reducing} and a class of other boosting \cite{deng2011fast,wang2012fast,paisitkriangkrai2014scalable} techniques have been proposed for lowering the testing complexity of machine-learning problems, they suffer from major limitations: a) In ensemble learning, a set of weak learners are combined into a complex classifier with high accuracy. The number of weak classifiers can be in the order of hundreds to get a reasonable performance for large-scale problems. Thus, ensemble methods become computationally expensive for larger datasets. b) Most existing models deviate from the biological attention based visual processing in the human brain and perform one-against-rest classification. In this case, the learning algorithm fails to maintain a general-to-specific feature hierarchy that turns out to be ineffective as well as computationally inefficient. 

A class of work on ``One-versus-All" and ``One-versus One" methods  have been explored to convert a multi-class problem into multiple binary classification problems. In such models, classes are not organized in a hierarchical tree. Also, these methods do not incorporate class relationships or feature similarities. An extension of these methods include \textit{error correcting output codes} \cite{allwein2000reducing} 
that utilize feature sharing to build more generalized and robust classifiers. As discussed earlier, these methods yield good classification accuracy. However, the time complexity is linearly proportional to the number of classes that does not scale well to larger datasets.

\cite{deng2011fast, griffin2008learning,gao2011discriminative} propose different ways to construct a hierarchical classification tree. However, most of these methods rely on a greedy prediction algorithm for class prediction through a single path of the tree. While these algorithms achieve sublinear complexity, the accuracy is typically sacrificed as errors made at higher nodes of the hierarchy cannot be corrected later. Researchers have also looked at developing efficient and effective feature representations for large-scale classification problems \cite{lin2011large,lowe2004distinctive, dalal2005histograms}.\cite{krizhevsky2012imagenet,he2015delving } learn discriminative features using deep convolutional networks to achieve state-of-the-art accuracy. Please note that our proposed ATree is orthogonal to such models since our method can use various feature respresentations to explore the accuracy vs. efficiency tradeoff. Hence, we do not optimize over different features in this work, rather compare the efficiency benefits of our approach with existing hierarchical methods. 

While our proposed ATree model draws inspiration from other tree-based methods such as \cite{grossmann2004adatree,tu2005probabilistic,gao2011discriminative}, we have different focus, design and evaluation strategies. As mentioned, most of these methods use a greedy prediction algorithm to achieve a good tradeoff between efficiency and complexity. The novelty of our work is that we use the recursive Adaboost training \cite{yuksel2012twenty,collins2002logistic} as a unified and principled optimization procedure to determine data partitioning (or learning attention hierarchy) based on feature similarity. This in turn enables the binary SVM to construct a maximum-margin hyperplane for optimal decision boundary modeling (with lower generalization error) leading to better performance. In addition, organizing the binary classifiers in a hierarchical tree structure on top of the attention hierarchy further reduces complexity. 

\section{Attention Model: Overview}
We use a variant of the boosted tree algorithm \cite{grossmann2004adatree,tu2005probabilistic} that combines Adaboost with a SVM based decision tree to construct the ATree. The proposed attention based classification framework resolves the problems associated with standard decision tree methods as discussed above. For a simple two-class (yes/no) problem, the training stage for ATree consists of two phases: a) First, we construct the visual feature hierarchy in ATree using the Adaboost training algorithm recursively, wherein each tree node is a complex classifier that works on an optimal feature at that tree level for partitioning the inputs. The partitioned input data obtained at a particular node are then used to train the left and the right sub-trees. 
Thus, the training data for the subsequent nodes of the tree are continuously pruned during the construction of ATree leading to computationally efficient training. The recursive boosting procedure intrinsically embeds clustering in the learning stage with similar feature clusters created in an automatic and hierarchical fashion. b) With the feature hierarchy and the resultant pruned data space fixed for each node/branch of the tree in the first phase, we train a standard binary SVM on the right and the left partitioned subsets of input data at each tree node. 

We further extend the two-class attention model described above to multi-class problems. 
We use the minimum entropy measure to select a feature that can be used to categorize the multiple category of objects into two broad classes. Then, the training algorithm for a simple two-class ATree is used to design the attention hierarchy. Again, clusters of multiple classes are automatically formed. At test time, only those branches and nodes depending upon the output of the binary classifier are activated that are relevant to the input. Hence, our approach is both time and energy efficient since it involves instance-specific selective activation of nodes.

Next, we briefly discuss the Adaboost learning framework and the shortcomings associated with it. We explain the intuition behind modifying the standard Adaboost training procedure which can then be effectively used to construct the feature-based hierarchy or ATree. 

\subsection*{Adaboost: Key Principles and Limitations}
The Adaboost algorithm combines a set of simple or weak classifiers ($h_t(x)$) \cite{schapire2013explaining} to form the final classifier ($H(x)$) given by $H(x)= \sum_{t=1}^{T}\alpha_t * h_t(x)$. The output of the final or strong classifier is $f(x)=sign(H(x))$. The weak classifiers can be thought of as feature or basis vectors. Given a set of training samples, Adaboost maintains a probability distribution, $W$ (uniform in the first iteration), of the samples. Then, Adaboost calls a WeakLearn algorithm \cite{schapire2013explaining} that trains the weak learner or classifier ($h_t$) on the weighted sample in a series of iterations \{$t=1,2,..T$\}. The distribution $W$ is updated in each iteration to minimize the overall error ($\epsilon$). Finally, Adaboost uses a weighted linear combination of the weak learners (or features) to obtain the final output $f$. The Adaboost and WeakLearn algorithm have been explained in detail in \cite{schapire2013explaining,li2005study}. 

For each sample $x_i$ with weight $w_i$, the error rate ($\epsilon= \sum_{i} w_i[sign(H(x_i)) \neq y_i$) is given by
\begin{equation}
\epsilon \le 2^T \prod_{t=1}^{T} \sqrt{\epsilon_t(1-\epsilon_t)}
\end{equation}
Eqn. 1 shows the maximum value of the error. For large-scale problems, when $x_i$ tends to be complex, $\epsilon_t$ saturates at $\frac{1}{2}$ after few iterations and thus the Adaboost algorithm fails to reach a global error minima. A possible solution for avoiding this is to design better weak classifiers that can effectively separate the classes. However, this would further increase the computational complexity for computing these classifiers (or features). 

One of the key principles of Adaboost is that “easy” samples that are correctly classified by the weak classifiers get low weights while those misclassified (“hard” samples) get higher weights. The weight distribution $W$ captures all the information about selected “features” in a given iteration. However, due to the weight update rule and normalization of W in each iteration, the information about previously selected features might be lost. This will result in misclassification of correctly classified (“easy”) samples from earlier iterations in the present epoch. Thus, the algorithm does not maintain a generic-to-specific transition while learning the weak classifiers (or features) that proves to be ineffective after a few iterations. To address this, we build an attention tree of strong classifiers, instead of constructing a single strong classifier from a linear combination of weak learners. The tree utilizes the features learnt from the previous nodes to construct the subsequent nodes. As we traverse down the tree, the classifiers learn more specific features that are useful for classifying the “hard” inputs correctly while preserving the feature information learnt at the early nodes for “easy” input samples. 

\section{Attention Tree: Learning and Implementation}
The central idea of the attention tree algorithm is to use feature-based attention to optimize the search procedure inherent in a vision problem. This model of attention addresses the reduction of number of candidate image subsets and feature subsets that are required for object recognition by selectively tuning the visual processing hierarchy. The theory described here is most closely related to the neuroscience works of \cite{tsotsos1995modeling} 
that present the neurobiological concepts of primate visual attention. 

\subsection{Training the attention tree for a 2-class problem}
Algorithm 1 shows the procedure for training the attention tree for a two-class problem. In \textit{Phase I}, a tree is recursively trained. It learns and preserves a hierarchy of features essential for understanding the underlying image representations and for efficient classification. At each node, a classifier is learnt using the Adaboost algorithm described in \cite{schapire2013explaining} that identifies the most optimum feature to separate the training inputs at a particular node into the corresponding sub-branches. It is shown in \cite{friedman2000additive} that Adaboost is essentially approximating a logistic regression. For convenience in notation, we denote the output computed by each classifier at the tree node as
\begin{align}
p(+1|x) &= \frac{1}{1+exp(-H(x))}\\
p(-1|x) &= \frac{1}{1+exp(H(x))}
\end{align}

\setlength{\textfloatsep}{0pt}
\begin{algorithm}[t]
\caption{2-class ATree}
\label{algo1}
\textbf{\textit{Phase I: Learning visual feature hierarchy}}
\vspace{1mm}
\newline \textbf{Input:} Training dataset D=$\{(x_1,y_1,w_1), …. , (x_n,y_n,w_n)\}$; $y_i \in \{+1,-1\}$, $\Sigma_iw_i=1$\\
 \textbf{Output:} ATree with feature hierarchy of depth L
\begin{algorithmic}[1]
\State \textbf{initialize} $tree_{depth}$=1
\State \textbf{while} ($tree_{depth}\le L$)
\State Using Adaboost, Train a strong classifier on D combining T weak classifiers.
\newline Calculate training error $\epsilon_t=\sum_{i=1}^{N} {w_i}^t$ , $y_i\neq h_t(x_i)$. EXIT Adaboost if $\epsilon_t> \gamma$ (user-defined, $\gamma$=0.48 in our experiments).
\newline{//\textit{Notations are same as explained in earlier section}}
\State Compute the probability distribution $\widehat{p}(y)=\Sigma_i w_i \delta(y_i=y)$
\State \textbf{initialize} $D_{left}$, $D_{right}$ = \{\}
\State \textbf{for} $i = 1:n$ //\textit{n=\# of samples}
\State Compute $p(+1|x_i)$ and $p(-1|x_i)$ using Eqn 2 and 3 for the strong classifier learnt in Step 3.
\State \textbf{if} $(p(+1|x_i) > \Delta)$ \textbf{then} $D_{right}={(x_i,y_i,1)}$, assign $y_i=\{+\}$
\State \textbf{elseif}  $(p(-1|x_i) > \Delta)$ \textbf{then} $D_{left}={(x_i,y_i,1)}$, assign $y_i=\{-\}$
\State \textbf{else} $D_{right}={(x_i,y_i, p(+1|x_i))}$ and $D_{left}={(x_i,y_i, p(-1|x_i))}$, assign $y_i=\{*\}$
\State \textbf{end if}
\State \textbf{end for}
\State $tree_{depth} ++$
\State Normalize weights in $D_{left}$ subset and goto Step 2. 
\State Normalize weights in $D_{right}$ subset and goto Step 2. 
\newline //\textit{Recursively repeat until $tree_{depth}$ is reached}
\State \textbf{end while}
\end{algorithmic}
\hspace{5mm}           - - - - - - - - - - - - - - - - - - - - - - - - - - - - - - 
\newline \textbf{\textit{Phase II: SVM optimization }}
\vspace{1mm}
\newline \textbf{Input:} ATree with feature hierarchy of depth L, Training samples ($D_{left}$ or $D_{right}$) for each node of the ATree=$\{(x_1,y_1), …. , (x_n,y_n); y_i \in \{+,-,*\}\}$\\
 \textbf{Output:} SVM based ATree
\begin{algorithmic}[1]
\State \textbf{initialize} $tree_{depth}$=1
\State \textbf{while} ($tree_{depth}\le L$)
\State Train a binary SVM at each tree node using $D_{left}$ and $D_{right}$ at that node ignoring all samples with $y_i=\{*\}$ with standard regularized hinge loss minimization \cite{li2005study}.
\State $tree_{depth} ++$
\State \textbf{end while}
\end{algorithmic}
\end{algorithm}

Depending upon the probabilities computed by the classifier node, the training set ($D$) is divided as $D_{left}$ and $D_{right}$ that are then passed to the sub-branches for training the following nodes of the tree. As the tree expands, only a subset of the input samples are passed to the subsequent nodes. Thus, the final nodes or leaves of the tree will consist of input samples belonging to one particular class. Please refer to Fig. 2 for an overview of the tree structure and input sub-sampling obtained with the attention model. Later, in section IV(D), we give a detailed explanation about the input sub-sampling and the hierarchical feature learning achieved with our attention model. 

In \textit{Phase II} of \textit{Algorithm 1}, a binary SVM (with any suitable kernel) is trained at each node of the tree using the $D_{right}$ and $D_{left}$ training sub-samples obtained from \textit{Phase I}. The training labels (+ for $D_{right}$, - for $D_{left}$ and * for instances that are passed to both $D_{left}$/$D_{right}$) are assigned to each input $x_i$ in the corresponding subsets for training the binary SVM. As the training set size decreases (owing to the input partitioning) at the successive nodes as we traverse down the tree, the complexity of the problem and hence that of the SVM also reduces. This in turn enables better decision boundary modeling with low computational complexity in the subsequent nodes (SVMs) for improved classification performance. Adding SVMs (\textit{Phase II} of Algorithm 1) at the nodes on top of the learnt feature hierarchy (\textit{Phase I} of Algorithm 1) enables the attention tree model to achieve state-of-the-art accuracies on challenging benchmark databases with significantly lower cost.

The threshold value, $\Delta$ in \textit{Phase I of Algorithm 1}, determines the fraction of training samples separated as positive (+) and negative (-) subsets. If  $\Delta$=1, then all training samples are passed to both branches (or sub-trees) of a tree node. The weights for both sub-trees are re-computed based on the node classifier’s output. In that case, the tree based Adaboost training converges to a standard boosting algorithm wherein the feature hierarchy (general-to-specific) is not learnt. For all our experiments discussed in section V, we set the $\Delta$ value to be $> 0.5$. For  $\Delta<0.5$, easy inputs that can be correctly classified with general features at the top nodes will be unnecessarily passed down to bottom nodes for classification. This will result in computational inefficiency, defeating the purpose of the attention tree. If  $\Delta=0.5$, then, each training sample is either passed to the right or left sub-tree which leads to a \textit{constrained} partition. In this case, the hard or confusing classes will be assigned to one of the sub-trees causing overfitting of data in the subsequent nodes. This will lead to a decline in accuracy. However, the test complexity will be low since the length of the tree will be short leading to a quicker decision at the cost of degraded performance. 

Those samples whose output probability lies in the range $[1- \Delta, \Delta]$ when $0.5<\Delta<1$ can be considered as hard or confusing ones. For $0.5<\Delta<1$, the hard samples are passed to both the left and the right sub-trees for training (*). The hard or confusing inputs/classes are ignored while training the SVM at the corresponding node in \textit{Phase II}. This is adopted from the \textit{relaxed hierarchy} structure in \cite{marszalek2008constructing,gao2011discriminative}. This is done to enhance the accuracy of the attention tree. It is understood that the decision boundary becomes progressively non-linear to model the hard or confusing classes in a dataset as we traverse down the ATree. The hard or confusing instances are ignored and passed to the bottom nodes that construct better decision boundary models, thereby, decreasing the overall error. In case the hard classes are not passed to bottom nodes, the SVMs at the top will construct overfitted models for the complex data instances, thereby, decreasing the accuracy considerably. In section V, we vary the threshold $\Delta$ to build \textit{constrained} and \textit{relaxed} hierarchical attention models and analyze the tradeoff between computational efficiency and accuracy for both approaches.

\subsection{Training the attention tree for multi-class problem}
To conserve the feature transition in the attention model, we propose a simple method for extending the two-class training model into a multi-class one. Traditionally, boosting algorithms use multi-class weak learners to construct a multi-class final strong classifier \cite{freunddecision,friedman2000additive}. However, for large number of classes, constructing reasonably accurate multi-class weak learners turns out to be highly computationally expensive. As seen earlier, we observe a feature similarity across classes that can be used to decompose a multi-class problem into a hierarchy of two-class problems.
 \setlength{\textfloatsep}{0pt}
\begin{algorithm}[t]
\caption{Multi-class ATree}
\label{algo1}
\textbf{Input:} Training dataset D=$\{(x_1,y_1,w_1),... , (x_m,y_m,w_m)\};$ $y_i \in \{1,...,n\}$, $\Sigma_iw_i=1$ \\
\textbf{Output:} SVM based ATree of depth L\\
For training a tree of maximum depth L,
\begin{algorithmic}[1]
\State Compute the probability distribution $\widehat{p}(y)=\Sigma_i w_i \delta(y_i=y)$
\State For each feature $f_j$ at value $v_j$, compute histogram $his_{left}(k)=\frac{1}{Z_{left}}\Sigma_i\delta(k=y_i)w_i$ for $y_i < v_j$ and $his_{right}(k)=\frac{1}{Z_{right}}\Sigma_i\delta(k=y_i)w_i$ for $y_i \ge v_j$.
\State Find optimal $f_j$ and $v_j$ that have minimum entropy $Z_{left}Entropy(his_{left})+Z_{right}Entropy(his_{right})$.
\State \textbf{if} $Z_{left}his_{left}(y_i) \ge Z_{right}his_{right}(y_i)$ \textbf{then} assign ${y_i}’=\{-1\}$
\State \textbf{else} assign ${y_i}’=\{+1\}$
\State \textbf{end if}
\State \textbf{initialize} training set D'=$\{(x_1,{y_1}',w_1), …. , (x_m,{y_m}',
 \newline w_m)\}$; $y_i \in \{-1, +1\}$
\newline //\textit{Now the multi-class is reduced to a 2-class problem}
\State Call Algorithm 1.
\end{algorithmic}
\end{algorithm}

Algorithm 2 shows the procedure for training a multi-class attention tree. The algorithm first finds the optimum feature across multiple classes that separates the input patterns into 2-classes and then uses the 2-class training procedure (\textit{Phase I} of Algorithm 1) to learn the subsequent classifier nodes of the tree. In our experiments, we observed that the feature chosen for transforming the multi-class to 2-class problem is often the feature selected by Algorithm 1 to construct the top node of the ATree. Intuitively, after the first selection, the features selected at the subsequent nodes help in making a stronger and more accurate decision.Thus, similar objects (with similar features) of different classes are clustered together in the initial nodes of the hierarchy. As the tree expands, these classes are gradually set apart. The tree is terminated when the algorithm does not find any common feature to partition the inputs (at the leaves of the tree). Thus, each leaf of the tree corresponds to a particular class. After the attention hierarchy is learned, \textit{Phase II} of Algorithm 1 is invoked to train SVMs at each node (excluding the leaves) of the hierarchy. 

\subsection{Testing the attention tree}
The attention tree composed of SVM nodes is then used for testing. Those instances (easy) that can be easily distinguished with general features are identified with SVMs at the top nodes. The SVMs at the bottom nodes perform more accurate classification on the hard instances in the dataset. When an input instance is presented at the root node, the branch with higher output probability at the SVM node is activated.  Based on the path activated by the output of SVM nodes, the instance then traverses the attention hierarchy until a leaf node where a final decision (or class assignment) is made. Note that a subset of classes are eliminated at each tree node as the tree is traversed. The attention based hierarchy, thus, scales sub-linearly \textit{O(log(n))} with respect to the number of classes. In the current era of “data deluge” that presents vision problems with a hefty task of recognizing from hundreds or thousands of classes, the sub-linearly growing attention tree model can be very useful. 

\subsection{Understanding the attention hierarchy}
\begin{figure}[!t]
\centering
\includegraphics[width=0.5\textwidth]{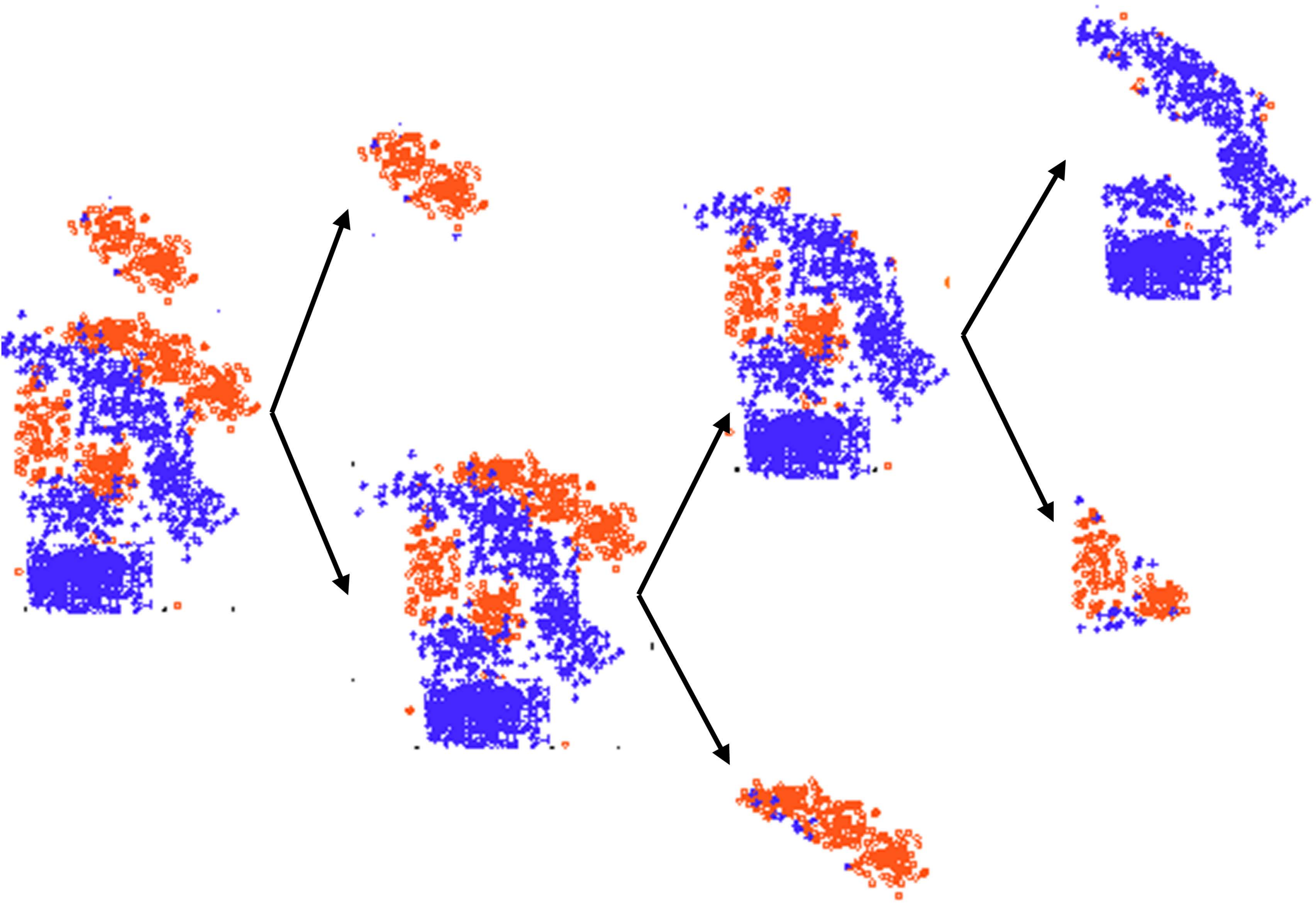}
\vspace{-4mm}
\caption{Attention Tree learning hierarchy formed for a synthesized dataset of 3000 points. Features for weak classifiers are position or distance to some specific 2D lines}
\end{figure}
The training algorithm naturally divides the samples into left and right sub-groups based on the configuration of features. Fig. 2 shows an example of how the attention tree learns and divides the samples on a synthesized dataset of 3000 points. The dataset consists of inputs belonging to two classes (denoted as orange and blue). The samples that are clustered together can be termed as hard inputs. Such samples are passed down the sub-branches of the tree forming the successive nodes. The top node of the tree partitions the inputs into two subsets. This division is intuitive as the right set of orange points are distant from the remaining inputs that are clustered together. The tree then expands on the hard inputs where the two sets are clustered together. If these data points are assumed to be features (like texture or color components) corresponding to two image classes, it is clearly seen that the hierarchy formed is coherent with the basic generic-to-specific feature transition theory of the attention model. 

Consider an example of recognizing a red Ferrari (blue points) from a sample set of vehicle images consisting of motorbikes and cars (orange points). The first intuitive step is to recognize all red vehicles in the sample and then look for a Ferrari shaped object from the sub-sample of remaining red images. The attention tree tries to model this intuitive behavior by learning the feature hierarchy. At the first level, the tree uses the red feature to distinguish the non-red vehicles from the red vehicles. As we go down, the tree uses more specific features (like Ferrari shapes or wheel textures) to perform more accurate classification. Our attention model automatically learns this feature hierarchy without any need to pre-specify the feature clusters. The pruning of the input data as we traverse down the attention model reduces the complexity of the original multi-class problem. This in turn enables better decision boundary modelling at the bottom SVM nodes of the ATree as compared to the top node resulting in improved classification performance. 

A noteworthy observation here is that the attention model comprises of multiple decision paths of different lengths. In Fig. 2, the tree consists of leaf nodes (for orange data points) at every level. For a given input, the decision can be reached at an earlier leaf node yielding a more optimal speedup during testing. Referring to the Ferrari example, all non-red images can be classified at the first level without traversing the whole tree. This imbalanced decision tree structure is what separates our model from other decision tree methods where one has to traverse the entire tree to reach a decision \cite{gao2011discriminative, tu2005probabilistic}. Even within a particular class, all inputs are not equal. For example, recognizing a person standing against a plain background is much easier (less time and effort) than when he/she is in the midst of a crowd. Ideally, algorithms should spend effort proportional to the difficulty of the inputs irrespective of whether they belong to the same class or not \cite{panda2016conditional}. Most existing works \cite{platt1999large,gao2011discriminative,marszalek2008constructing} focus on optimizing the computational complexity based on inter-class feature variability. In contrast, our imbalanced method captures both inter and intra class feature variability while expanding the attention tree thus yielding more computational benefits.

\subsection{Constrained vs Relaxed Hierarchy}
\begin{figure}[!t]
\centering
\includegraphics[width=0.5\textwidth]{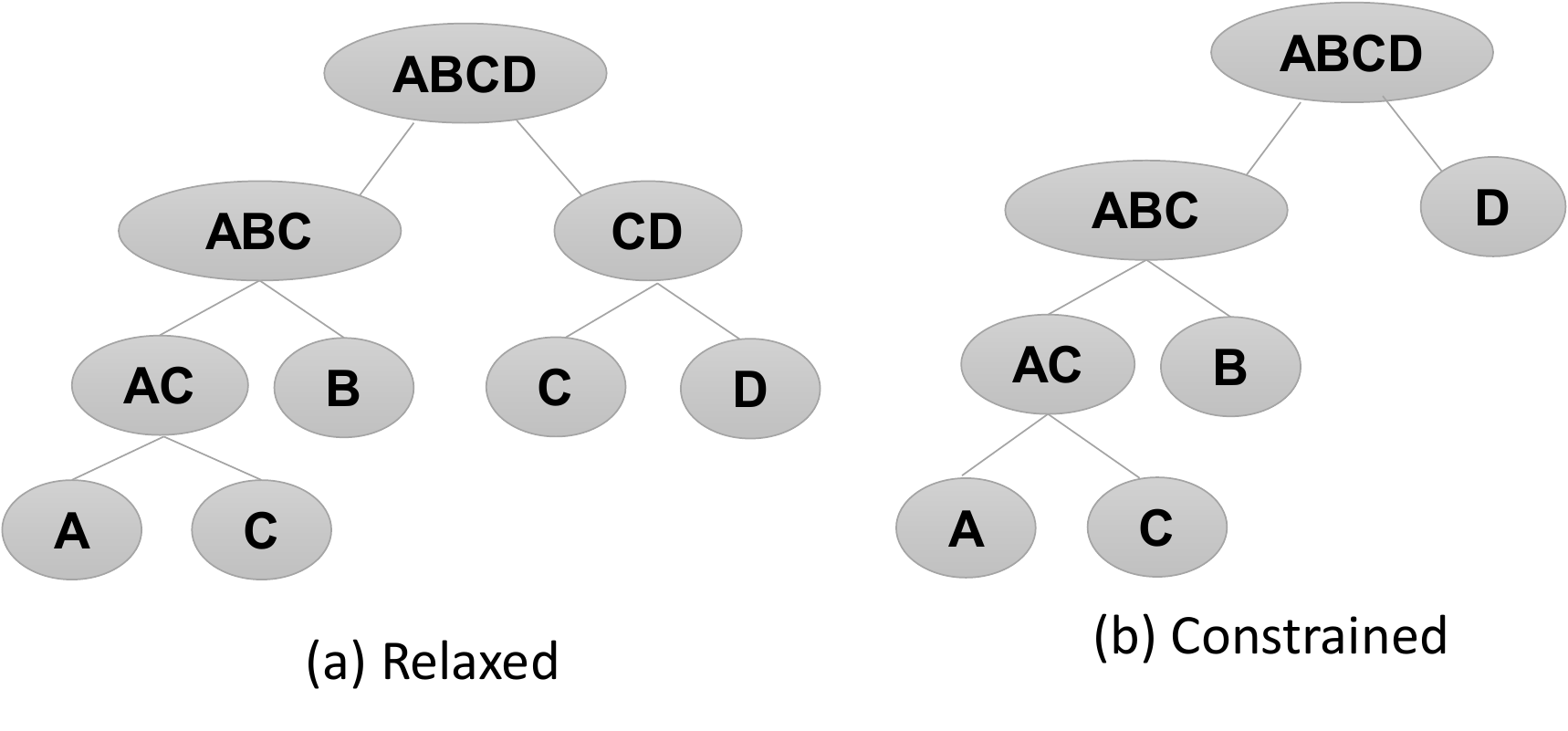}
\vspace{-4mm}
\caption{Sample illustration of ATree with Relaxed and Constrained hierarchy}
\end{figure}
Previously, we discussed that the threshold, $\Delta$, serves as a useful control parameter to construct either relaxed or constrained models of the attention tree. Fig. 3 demonstrates the sample relaxed/constrained hierarchy for a 4-class problem. The instances from class C are the hard inputs in the dataset. In the constrained hierarchy, it is clearly seen that instances from C are forced to the left sub-node. In this case, it is very likely that the SVM at the root node will misclassify a test instance from class C due to overfitting. However, the decision path for recognizing class D is short. So, we will observe an improvement in efficiency (or test speed) at the cost of accuracy. With relaxed hierarchy, we see that there is an extra SVM classifier evaluation required to recognize class D that increases the computational cost. However, the accuracy in this case will be better as the addition of an extra classifier node (Node CD in Fig. 3(a)) minimizes overfitting for complex distribution of data. In addition, the relaxed hierarchy captures the intra-class feature variability for class C which is not seen in the constrained model. In the relaxed model, instances of class C that are relatively easy can be classified at the 2nd level and those that are hard are only passed to the 3rd level for accurate classification. In contrast, with constrained model all instances of class C are passed to the 3rd level for classification. From this sample demonstration, it is clear that $\Delta$ can be modulated to control the accuracy and efficiency of the ATree. 

\subsection{Optimizing the computational cost for SVMs with non-linear kernels}
Since the learnt tree with SVM nodes is used for measuring the complexity and accuracy of the attention model, the kernel-type selection (non-linear or linear) plays a key role in determining the overall computational efficiency/performance for a given multi-class problem. In case of SVMs with linear kernels, the complexity of each classifier node is same, so the overall test complexity is proportional to the number of classifiers evaluated to reach a decision. However, in case of non-linear kernels, the complexity of each classifier is proportional to the number of its support vectors. So, we use a computational model as devised in \cite{gao2011discriminative} to optimize the number of support vectors for maximizing the computational benefits. It is clear that the training algorithm for multi-class attention tree does not always result in a balanced partitioning of classes (into left and right sub-trees) at a particular node as observed in Fig. 3. Given a SVM classifier $M$, let $N(M)$ be the number of support vectors of the classifier $M$. We define a cost function ($c(M)$) that reflects the average efficiency of the SVM ($M$) as:
\begin{equation}
c(M)=f^-(M)\frac{N(M)}{|Z^+(M)|} + f^+(M)\frac{N(M)}{|Z^-(M)|}
\end{equation}
where $|Z^+|$ and $|Z^-|$ are the number of classes assigned to positive (right sub-tree) and negative (left sub-tree) labels respectively. $f^-(M) = \frac{|Z^-(M)|}{|Z^+(M)|+|Z^-(M)|}$ is the fraction of negative classes and similarly $f^+(M)$ is the fraction of positive classes. After the attention hierarchy is learned (\textit{Phase I} of \textit{Algorithm1}), we can estimate $|Z^+|$ and $|Z^-|$. In an ideal case, for instances from class $|Z^-|$, classes $|Z^+|$ are pruned after evaluating $M$ with cost proportional to $N$ number of kernel evaluations. So, the average cost for discarding a particular class is $\frac{N(M)}{|Z^+(M)|}$. Similarly, the average cost for eliminating a class for instances belonging to positive subset ($|Z^+|$) is $\frac{N(M)}{|Z^-(M)|}$. Given the proportion of positive $f^+(M)$ and negative $f^-(M)$ classes, the average cost for eliminating one class by $M$ is given by Eqn. 4. Thus, we select the number of support vectors, $N(M)$, that minimizes the overall cost function $c(M)$ while yielding competitive accuracy. 

\section{Experiments}
In this section, we evaluate our proposed framework on two fundamental computer vision tasks: object recognition and scene categorization for the benchmark datasets, Caltech-256 \cite{griffin2007caltech} and SUN \cite{xiao2010sun}. 

\subsection{Basic Setup}
We use the evaluation metrics: classification accuracy and test speed (or test complexity) to discuss the benefits of our approach. For classification accuracy, we use the mean of per-class accuracy that is reported as a standard way for estimating multi-class classification performance. For test speed, we distinguish two cases based on the kernel type selection for the SVM classifiers at the ATree nodes. The first case corresponds to linear classifiers, where the overall test complexity is proportional to the number of evaluated classifiers. So, for a linear kernel SVM, we report the mean of the number of classifier evaluations for all test instances. The second case corresponds to nonlinear kernel SVMs. As mentioned earlier, the complexity of each classifier is now proportional to the number of support vectors. Specifically, let $M$ be the number of classifiers to be evaluated, where each classifier has a set of support vectors $\{SV_i^{(m)}\}$ where m and i denote the classifier and its support vectors respectively. If $M$ classifiers are evaluated independently without caching any kernel computations, then, the number of kernel computations for a single test instance is given by $\sum_{m=1}^{M} |\{SV_i^{(m)}\}|$. This method proves to be very inefficient when the number of classifiers are large. An efficient approach would be caching the kernel computations from different classifiers and reusing them whenever possible. Then, the number of kernel computations reduces to $|\cup_{m=1}^{ M}\{SV_i^{(m)}\}|$. We use the latter approach to report test speed when non-linear kernels are used.

We compare our method to various existing approaches: Gao \cite{gao2011discriminative}, one-vs-all, one-vs-one, DAGSVM, tree-based hierarchy \cite{griffin2008learning} and Marszalek \cite{marszalek2008constructing}. The regularization parameter $C$ of SVM is chosen by cross validation on the training set.

\begin{figure*}[!t]
\centering
\subfloat[Linear Kernel SVM]{\includegraphics[scale=0.37]{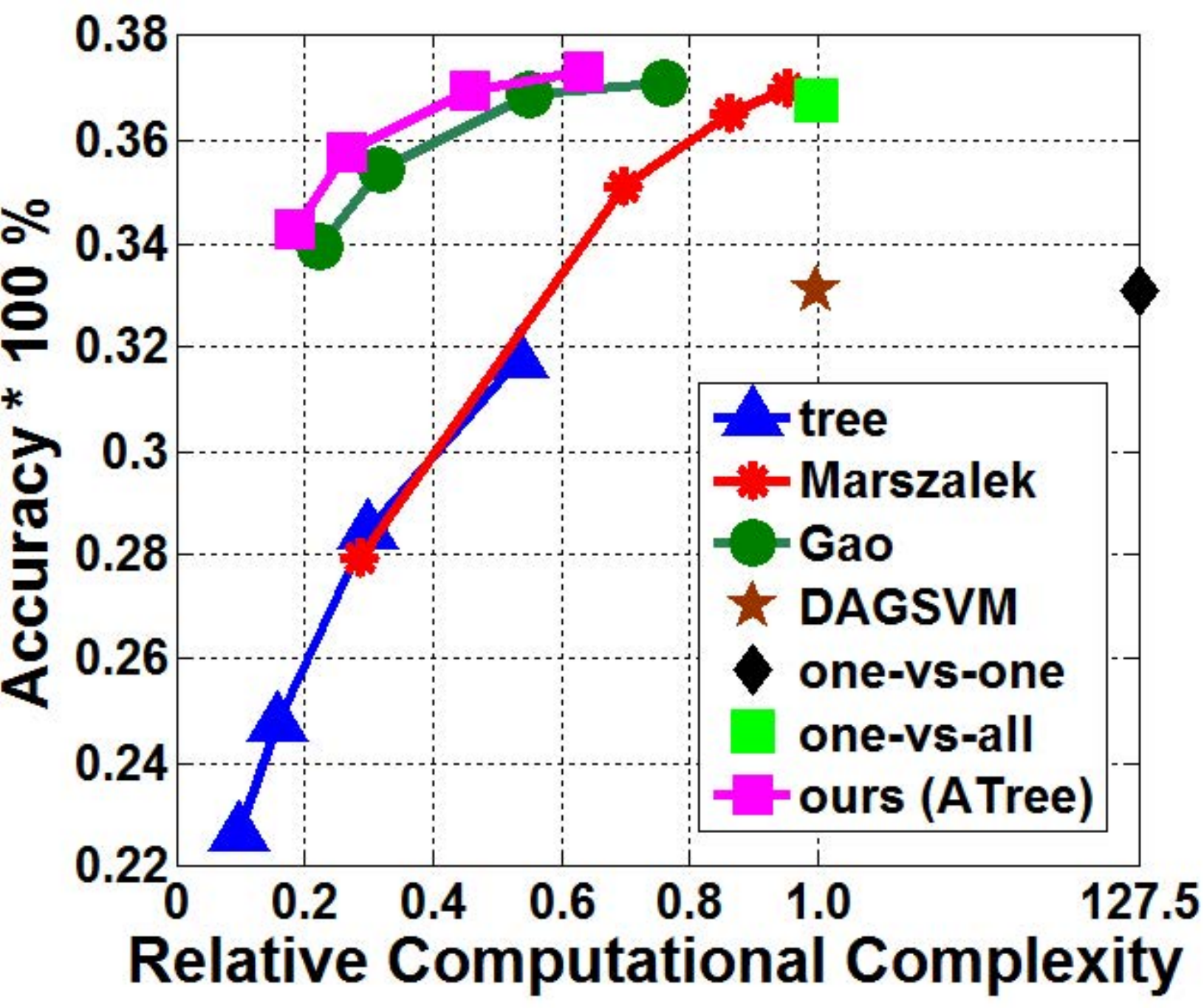}
}
\hspace{5mm}
\subfloat[Gaussian kernel SVM based on $\chi^2$ distance]{\includegraphics[scale=0.37]{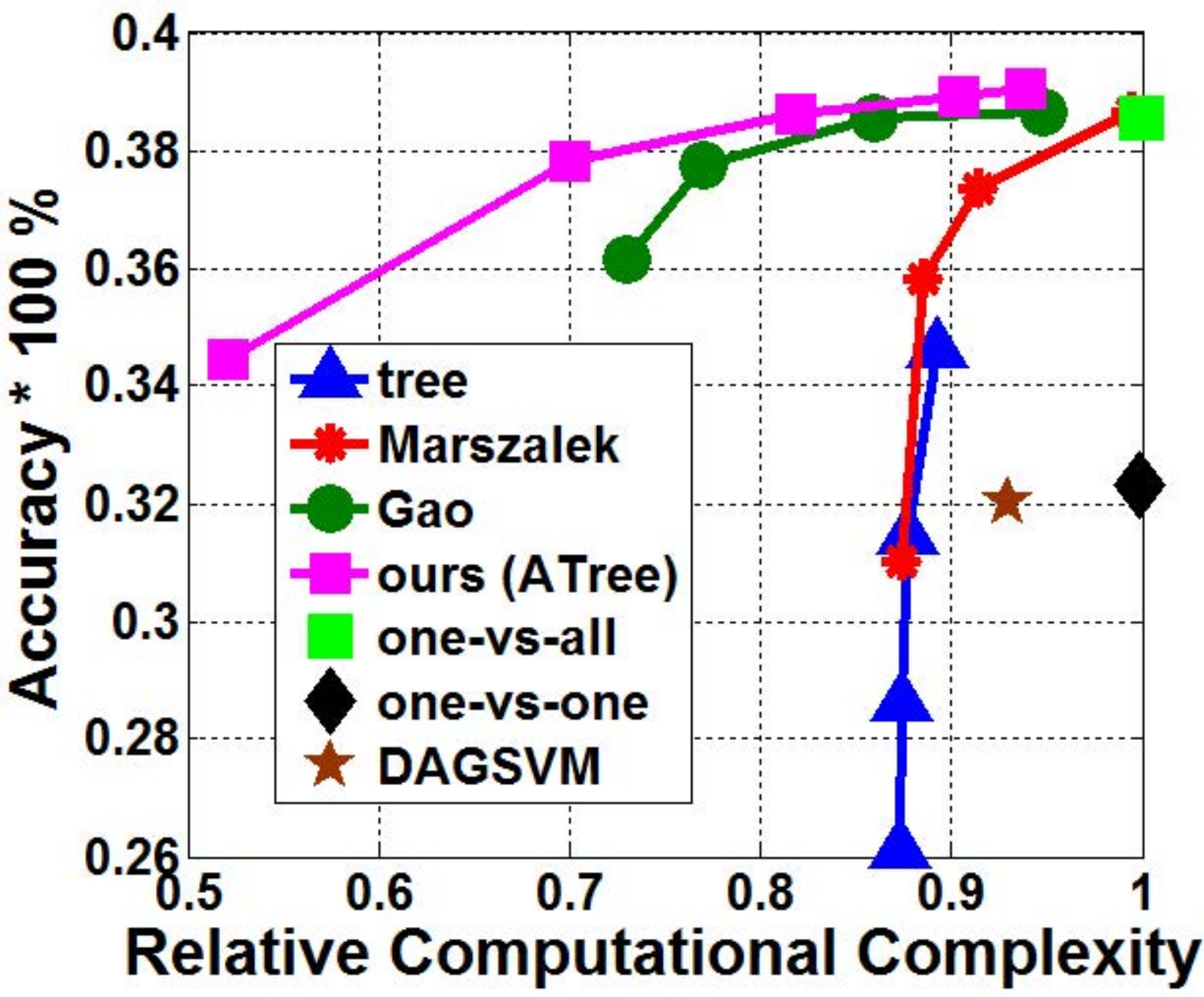}
}
\caption{Comparison of the tradeoff between accuracy and Relative Complexity on Caltech 256. The computational complexity is normalized by the complexity of one-vs-all. Note that for one-vs-one the relative complexity with linear kernel is 127.5.}
\label{fig_sim}
\end{figure*}

\subsection{Caltech-256}
With 256 categories and at least 80 images per class, this is a standard muti-class object recognition dataset. We randomly sampled 80 images for each class, and used half (40 per class) for training and remaining half for testing. For features, we used the standard spatial histograms of visual words based on dense SIFT \cite{lowe2004distinctive}. Like \cite{gao2011discriminative}, we used the extended Gaussian kernel based on $\chi^2$ distance. However, since linear kernel of histogram based features gives poor accuracy, we used explicit feature transformation from \cite{vedaldi2012efficient} to approximate implicit feature mapping of  $\chi^2$ kernel. The linear SVM is applied on the transformed feature. We varied computational parameters for tree \cite{griffin2008learning} (2 to 5 levels), Marszalek \cite{marszalek2008constructing} ($\alpha \in \{0.2, 0.5, 0.6, 0.8\}$ ), Gao \cite{gao2011discriminative} ($\rho \in \{$0.5 to 0.8 with step size of 0.1\})  and our method ATree ($\Delta \in \{$0.5 to 0.9 in steps of 0.1\})  to obtain a tradeoff between accuracy and speed. Here, $\alpha$ and $\rho$ are the computational parameters defined in  \cite{marszalek2008constructing} and \cite{gao2011discriminative} respectively that are varied to achieve the complexity vs. accuracy tradeoff.

Fig. 4 shows the results. It is clearly seen that ATree performs better (faster at same accuracy and more accurate at the same Relative Complexity (RC)) for both linear and non-linear kernels. For instance, in case of linear kernel, ATree achieves one of the best accuracy ($\sim$37.3\%) with around 27\% of the complexity of one-vs-all with a \textit{relaxed hierarchical} model (where $0.5<\Delta<1$) while achieving a speedup of 3.7x. Also, for $\Delta=0.5$, when the ATree is modelled as a \textit{constrained hierarchy}, it achieves a higher speed up of 5.5x for $\sim$2.5\% accuracy degradation with respect to one-vs-all. However, to achieve a similar 5x speed up other methods: Gao \cite{gao2011discriminative}, Marszalek \cite{marszalek2008constructing}, tree  \cite{griffin2008learning} have to suffer 3.2\%, 8\%, 10\% accuracy degradation. Please note that ATree achieves consistently better accuracy performance than the best result reported in \cite{gao2011discriminative} for both linear and non-linear kernels.

\subsection{SUN}
Now, we evaluate our ATree model for scene classification on the SUN dataset. The SUN dataset captures a full variety of 899 scene categories. We used 397 well-sampled categories as \cite{xiao2010sun, gao2011discriminative}. For each class, 50 images are used for training and the other for test. For image representation, we used spatial HOG pyramid \cite{dalal2005histograms} with histogram intersection kernel (non-linear SVM) and transformed spatial Histogram Oriented Gradient (HOG) pyramid (explicit feature transformation from \cite{vedaldi2012efficient} to approximate the implicit feature mapping of histogram intersection kernel) with linear kernel (linear SVM). As with Caltech-256, we varied the tradeoff between accuracy and speed for tree \cite{griffin2008learning} (2 to 5 levels), Marszalek \cite{marszalek2008constructing} ($\alpha \in\{0.1, 0.2, 0.3, 0.4, 0.6, 0.8\}$), Gao \cite{gao2011discriminative} ($\rho \in \{$0.6 to 0.9 with step size of 0.1\}) and our method ATree ($\Delta \in \{$0.5 to 0.9 with step size of 0.05\}).
\begin{figure*}[!t]
\centering
\subfloat[Linear Kernel SVM]{\includegraphics[scale=0.37]{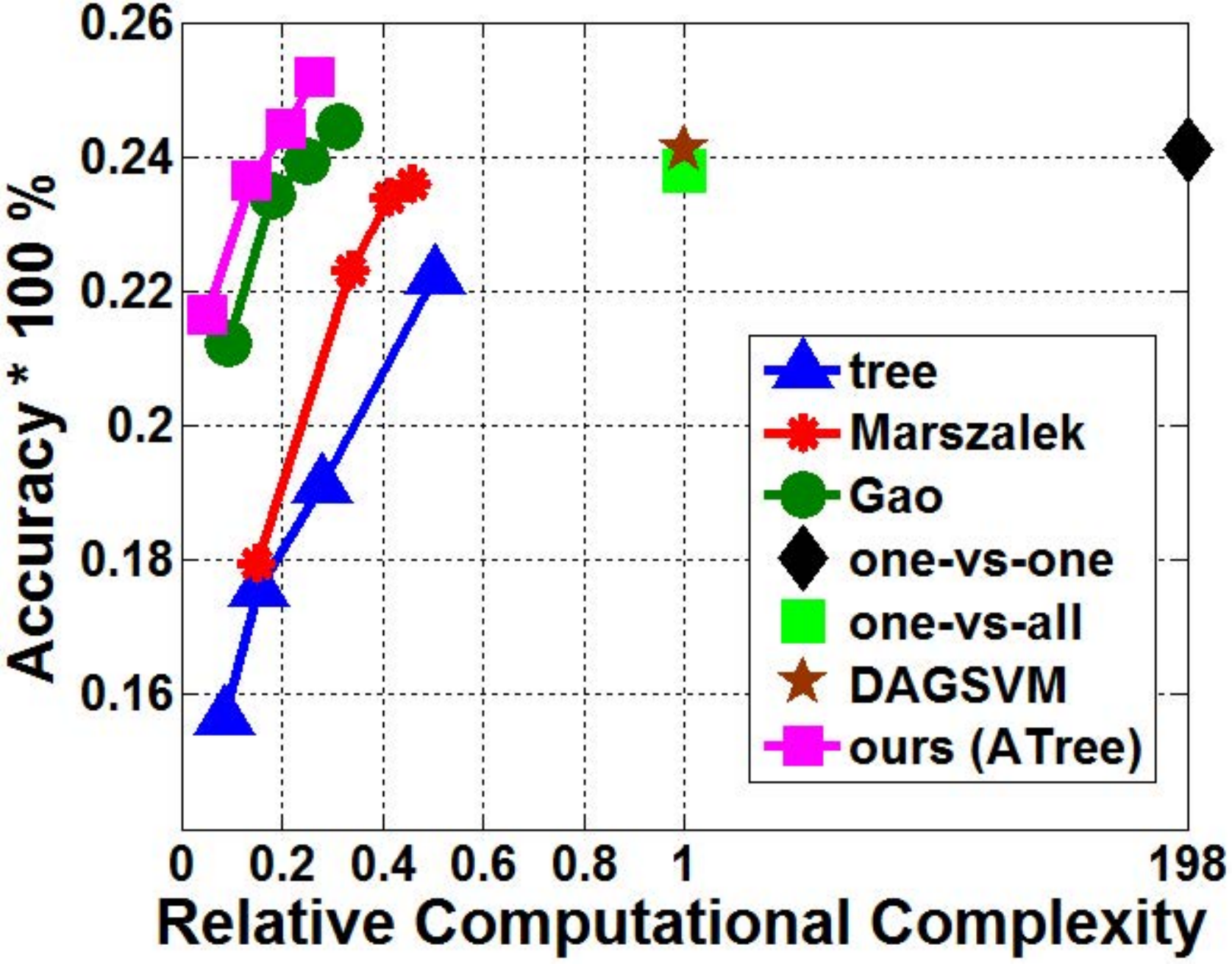}
}
\hspace{5mm}
\subfloat[Spatial HOG feature with histogram intersected kernel (non-linear SVM)]{\includegraphics[scale=0.37]{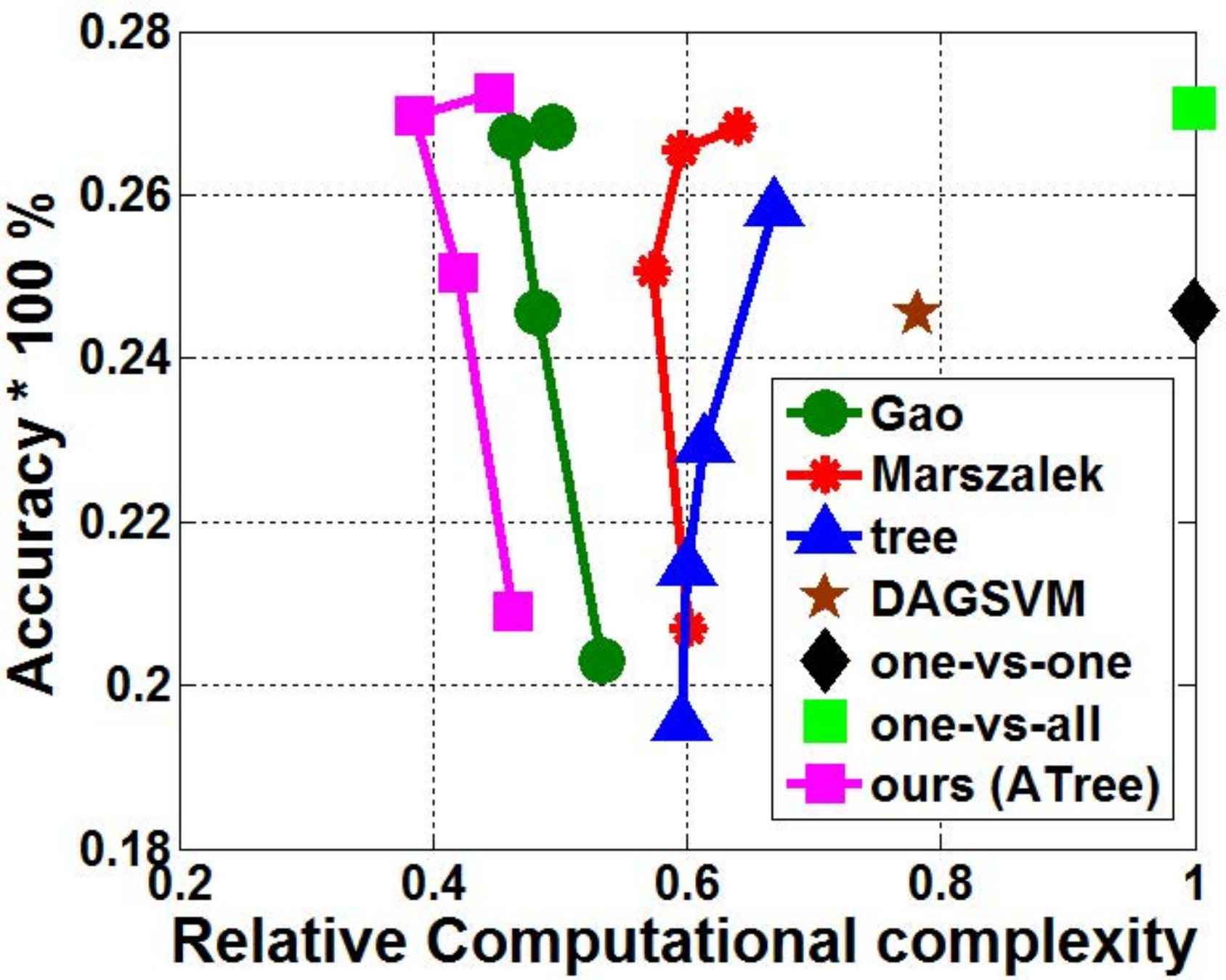}
}
\caption{Comparison of the tradeoff between accuracy and Relative Complexity on SUN. The computational complexity is normalized by the complexity of one-vs-all. Note that for one-vs-one the relative complexity with linear kernel is 198.}
\label{fig_sim}
\end{figure*}

Fig. 5 shows the results. The performance improvement for both linear/non-linear kernels is similar and consistent with the results of Caltech-256. For instance, for HOG with histogram intersection kernel, our method has a significantly improved accuracy of 24.4\% with $\sim$21\% complexity compared to one-vs-all ( for $\Delta$ $\sim$0.65 implying a \textit{relaxed hierarchy}). However, Marszalek and Gao can only reduce the relative complexity to 49\% and 64\% respectively to attain similar accuracy as one-vs-all. The performance of the ATree further improves if $\Delta$ is increased and the highest accuracy observed is 25.2\% (at 26\% complexity) that is $\sim$1.7\% higher than the best result reported in \cite{gao2011discriminative}. As for the test speed, while our method achieves a maximum speed up of 4.8x compared to one-vs-all even with an improved accuracy, other methods never meet this speedup irrespective of the accuracy. In addition, with linear kernel, ATree achieves a slightly improved accuracy with respect to one-vs-all and DAGSVM while being 7.2x faster. In fact, for a 1.8\% decline in accuracy, ATree (with \textit{constrained hierarchy} for $\Delta=0.5$) is 19x faster than one-vs-all. However, for Gao \cite{gao2011discriminative}/Marszalek \cite{marszalek2008constructing}, the accuracy degradation is higher upto 2.5\%/6.8\% to achieve similar speed up.  The above results validate that ATree is more effective to reduce the RC while maintaining a competitive accuracy in comparison to other hierarchical tree-based implementations.

From Fig. 5 (b), it is worth noting that if non-linear kernels are used, a lower depth tree does not necessarily lead to lower computational complexity. When $\alpha$ is large for \cite{marszalek2008constructing} or $\Delta$ is closer to 0.5 ($0.5< \Delta<0.6$), the depth of the tree is low on account of constrained partitioning of inputs into left and right sub-trees. Ideally, we should get an accuracy decline with a lower complexity for such cases as the number of classifier evaluations will be less. However, we observe that both accuracy and complexity are worse. The reason is that, although a fewer number of classifier evaluations are required in these cases, each SVM involves a large number of support vectors (since constrained partition forces the SVMs to perform complex boundary modelling) which increases the overall complexity.

Besides performance comparison, we also studied how the complexity of ATree changes with the increase in the number of classes. We sampled 100, 200 and 300 with the original 397 classes from SUN dataset, and for each case we learn the model with spatial HOG using linear kernel. For fair comparison, we set $\Delta=0.6$ for our method, $\alpha =0.4$ for \cite{marszalek2008constructing} and $\rho =0.7$ for \cite{gao2011discriminative} to match the same level of accuracy as one-vs-all. As seen in Fig. 6, the complexity of our method grows sublinearly as compared to \cite{marszalek2008constructing}. As discussed earlier, ATree model gives rise to an imbalanced tree that can have leaf nodes even at the beginning of the attention hierarchy. Thus, we observe that our method grows at a slightly lesser rate than that of \cite{gao2011discriminative}.

\subsection{Attention hierarchy of visual features}
ATree builds a feature hierarchy in the label space automatically. Fig. 7 shows the attention tree formed for a subset of some sampled images from the Caltech-256 dataset. We observe that the images that have similar features are clustered together in an initial node and are gradually set apart as the tree is traversed. Conforming to the imbalanced attention model, we observe that for certain classes: zebra, car tire, the classification is done at earlier nodes while more confusing classes are passed down. In addition, we also observe intra-class variability for the camel class in which certain instances are evaluated earlier than others. 

\begin{figure}[!t]
\centering
\includegraphics[width=0.35\textwidth]{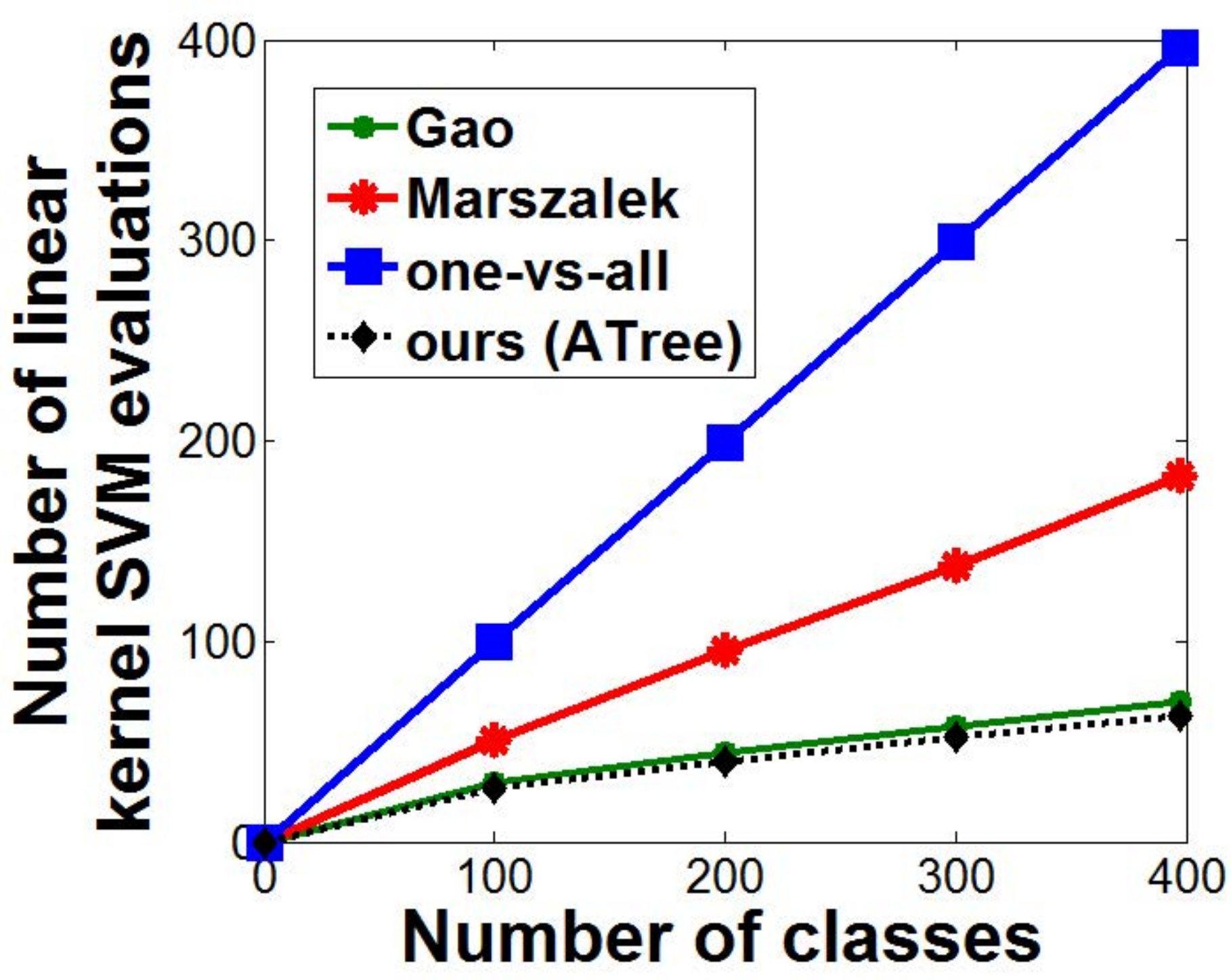}
\vspace{-4mm}
\caption{Sublinear growth in complexity with number of classes}
\end{figure}

\begin{figure}[!t]
\centering
\includegraphics[width=0.5\textwidth]{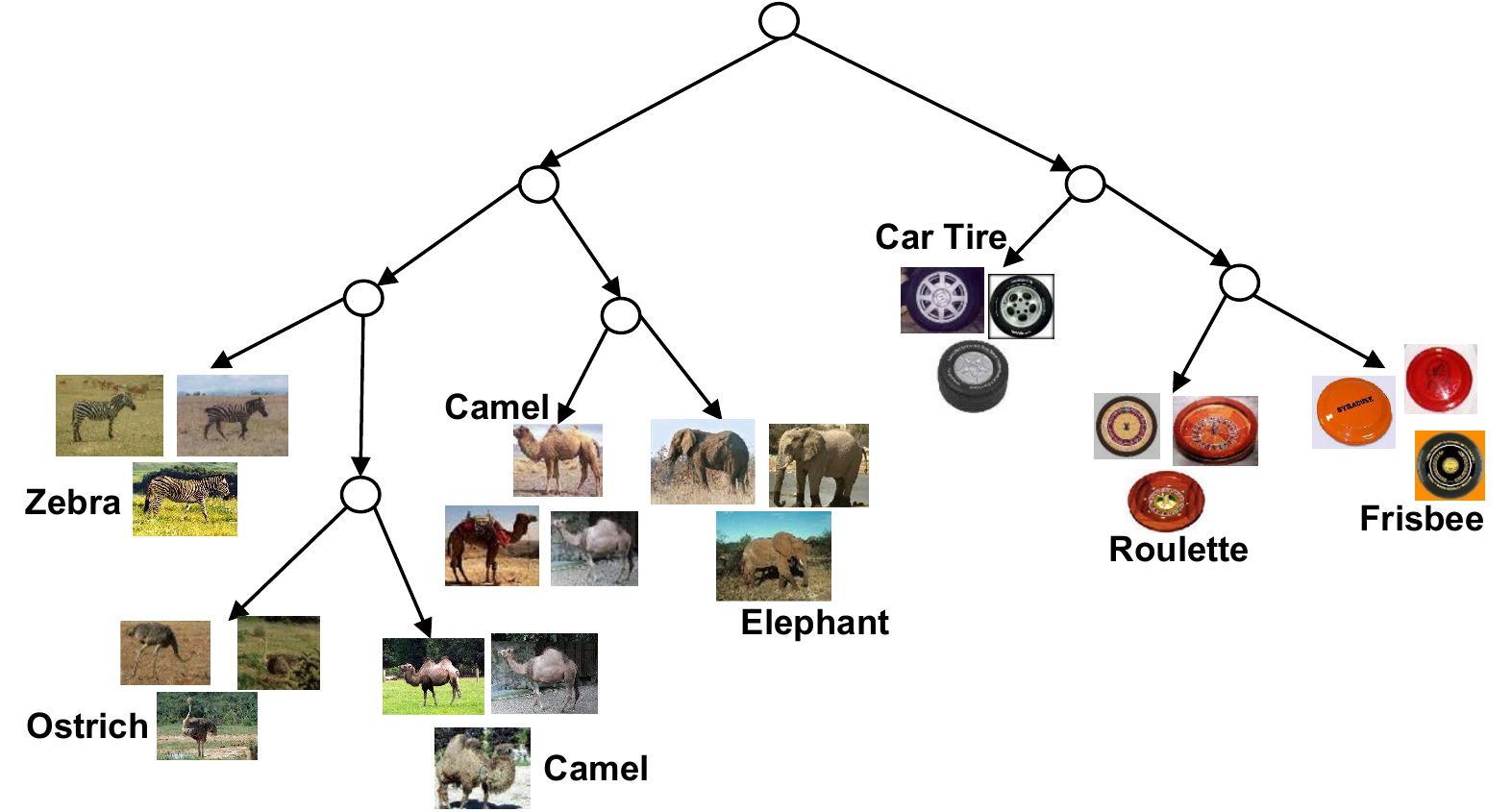}
\vspace{-4mm}
\caption{ATree formed for a sub-sample of selected images from Caltech-256}
\end{figure}

In Fig. 8, we present a sampling of the first three levels of the ATree constructed for the entire Caltech-256 and SUN dataset showing how the different classes are assigned \{+,-,* (Algorithm 1)\} and separated into left and right sub-trees. In the Caltech-256 ATree hierarchy, we observe that the assignment of classes into sub-nodes in many cases correlates to human vision i.e. images from different classes that are assigned to the same sub-tree 
look similar to humans. For the SUN ATree hierarchy, the partitioning of classes in the first two levels correlates with human-defined concepts. e.g. , natural outdoor scenes vs. indoor man-made scenes. Also, the hierarchy starts partitioning classes with large visual distances and then identifies subtle discrepancies at the bottom nodes which is in coherence with the concepts of visual stimuli decomposition in the human brain. This suggests the biological plausibility and effectiveness of our attention model for image classification.

\begin{figure*}[!t]
\centering
\subfloat[ATree using Gaussian kernel on Caltech-256]{\includegraphics[scale=0.75]{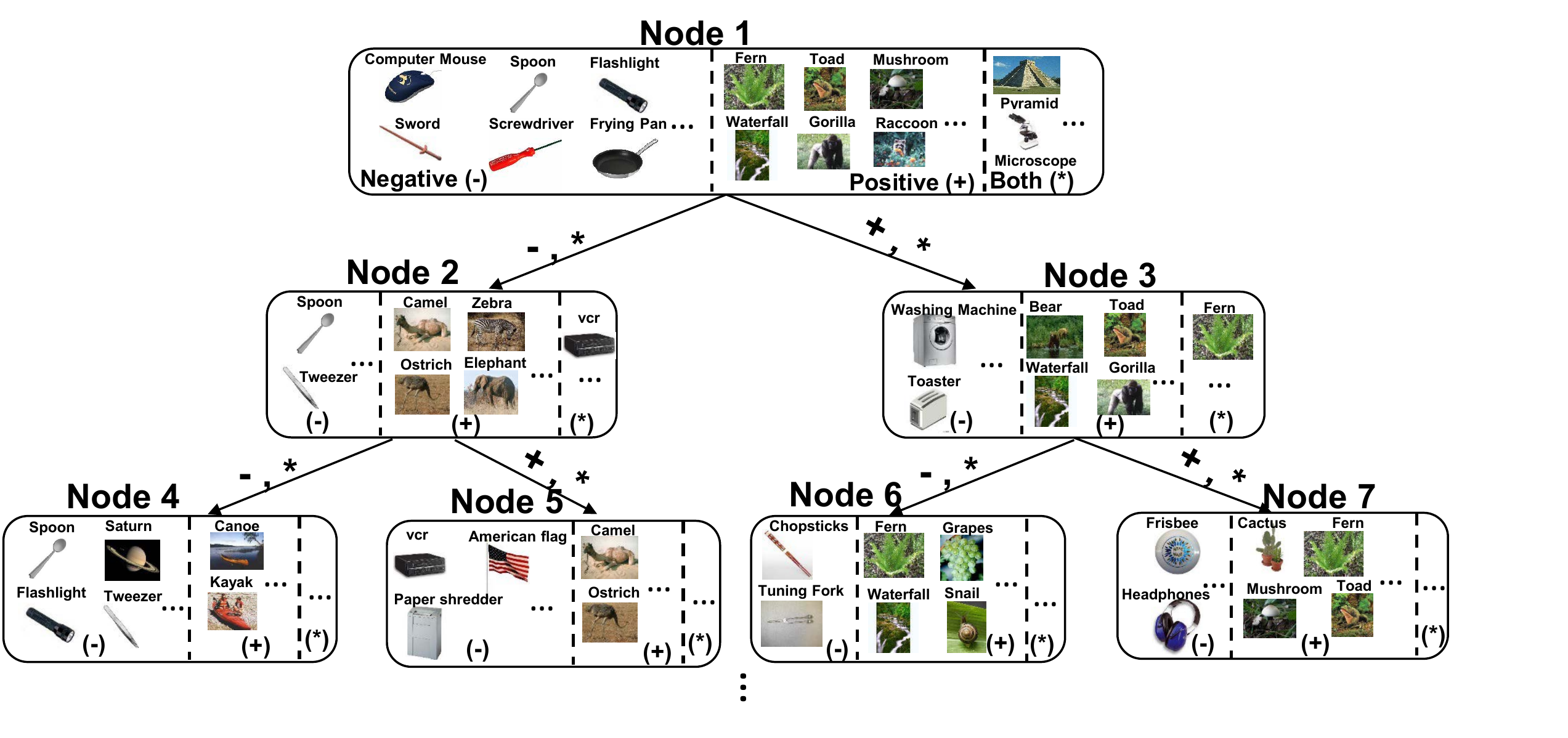}
}
\hfill
\subfloat[ATree using HOG on SUN]{\includegraphics[scale=0.75]{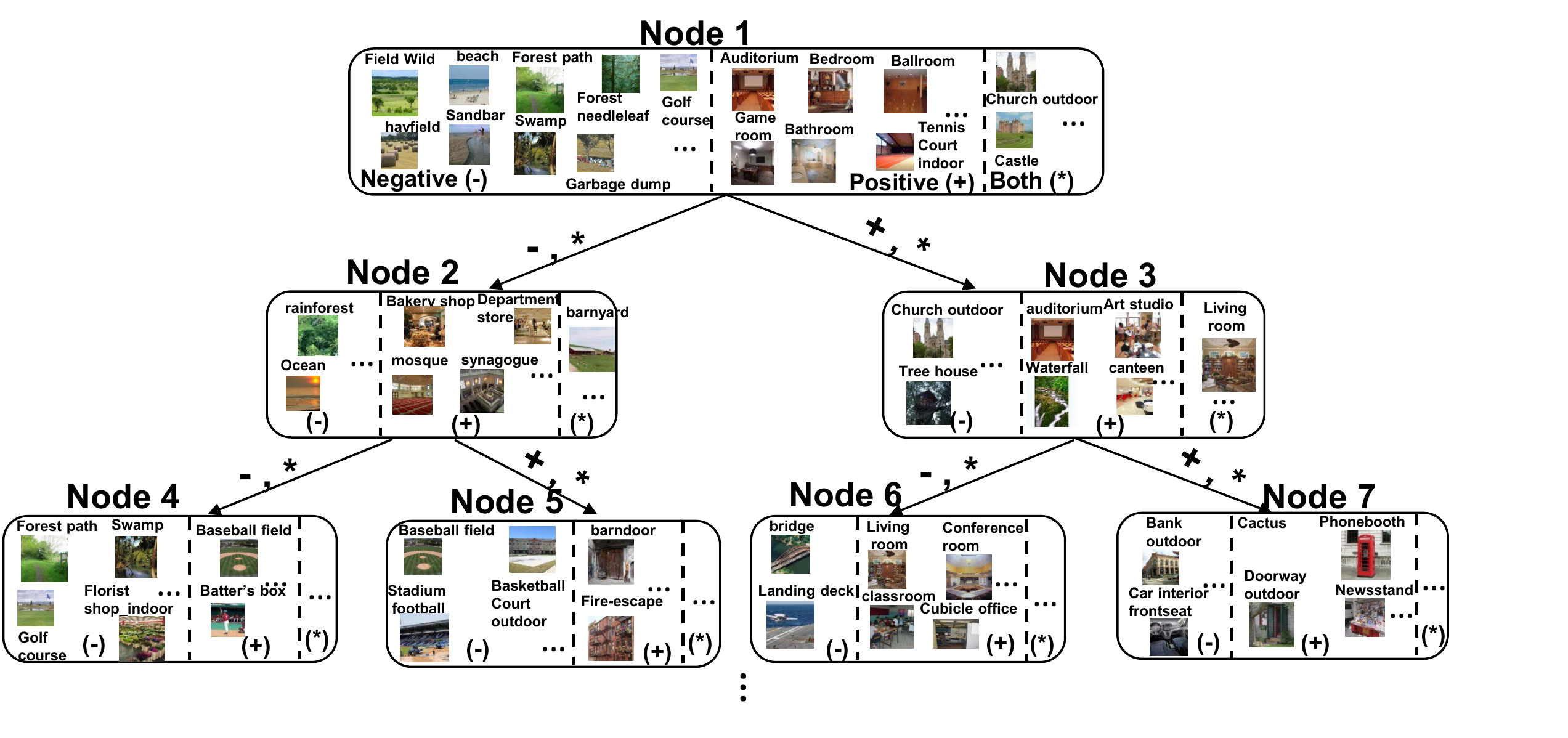}
}
\caption{First 3 levels of the ATree showing the data partitioning for a few examples at each node for both Caltech-256 and SUN dataset.}
\label{fig_sim}
\end{figure*}

\section{Discussion and Conclusion}
We proposed a novel neuro-inspired visual feature learning to construct an efficient and accurate tree-based classifier: Attention Tree, for large-scale image classification. Our learning algorithm is based on the biological attention mechanism observed in the brain that selects specific features for greater neural representations. The ATree uses a principled optimization procedure (recursive Adaboost training) to extract knowledge about the relationships between object types and integrates that into the visual appearance learning. We evaluated our method on both the Caltech-256 and SUN datasets and obtained significant improvement in accuracy and efficiency. In fact, ATree outperforms the one-vs-all method in accuracy and yields lower computational complexity compared to the state-of-the-art “tree-based”methods \cite{gao2011discriminative,marszalek2008constructing}. The proposed framework intrinsically embeds clustering in the learning procedure and identifies both inter and intra class variability. Most importantly, our proposed ATree learns the hierarchy in a systematic and less greedy way that grows sublinearly with the number of classes and hence proves to be very effective for large-scale classification problems. It is noteworthy to mention that the current ATree framework suffers from overfitting when the training dataset is small. The overfitting behaviour is checked by modulating the depth of the ATree and also adopting the relaxed hierarchy structure where confusing or “hard” inputs are passed to both the right and the left sub-nodes. Additionally, tree pruning methods \cite{kearns1998fast, yildiz2015vc} can be used to control overfitting . Further research can be done to explore the overfitting problem.

%
%
%
%
%
%
%
\section*{Acknowledgment}
This work was supported in part by C-SPIN, one of the six centers of StarNet, a Semiconductor Research Corporation Program, sponsored by MARCO and DARPA, by the Semiconductor Research Corporation, the National Science Foundation, Intel Corporation and by the National Security Science and Engineering Faculty Fellowship.

\ifCLASSOPTIONcaptionsoff
  \newpage
\fi

\end{document}